\documentclass{article}

\usepackage{iclr2018_conference,times}
\usepackage{url}

\usepackage{amsmath}
\usepackage{amsfonts}
\usepackage{amssymb}
\usepackage{mathtools}
\usepackage{graphicx}
\usepackage{hyperref}
\usepackage{subcaption}

\newcommand{\squeeze}{\vspace{-2.5mm}}

\title{On Periodic Functions as Regularizers \\ for Quantization of Neural Networks}

\author{Maxim Naumov, Utku Diril, Jongsoo Park, \\ \textbf{Benjamin Ray, Jedrzej Jablonski and Andrew Tulloch} \\ Facebook, 1 Hacker Way, Menlo Park, CA, 94025}

\begin{document}

\maketitle

\begin{abstract}
Deep learning models have been successfully used in computer vision and many other fields. We propose an unorthodox algorithm for performing quantization of the model parameters. In contrast with popular quantization schemes based on thresholds, we use a novel technique based on periodic functions, such as continuous trigonometric sine or cosine as well as non-continuous hat functions. We apply these functions component-wise and add the sum over the model parameters as a regularizer to the model loss during training. The frequency and amplitude hyper-parameters of these functions can be adjusted during training. The regularization pushes the weights into discrete points that can be encoded as integers. We show that using this technique the resulting quantized models exhibit the same accuracy as the original ones on CIFAR-10 and ImageNet datasets.
\end{abstract}

\section{Introduction}

Deep learning models require a very large amount of resources during their training (repeated forward and backward propagation) as well as inference (forward propagation). Further, the latter is often performed on the edge devices, such as smartphones or embedded systems, which operate within strict size, temperature and power budget \cite{Shimpi2011,Humrick2017,Dolbeau2018,Turing2018}. As a result these devices can perform a limited \# of operations per second\footnote{The peak for ARM Cortex is based on fp32 ``VMLA.F32 Qd, Qn Dm", fp16 ``VMLA.F16 Qd, Qn Dm" and int8 ``VMLAL.S8 Qd, Dn, Dm" instructions, with estimated reciprocal throughput $4,\dagger,2$ and width $4,\dagger,8$ for A-7, reciprocal throughput $1,1,1$ and width $4,8,8$ for A-75. Then, peak ops are defined as (frequency/throughput)*width*cores. Also, the power is assumed to be 750mW and 1W per core for ARM Cortex A-7 and A-75, respectively.}, as illustrated in Tab. \ref{tab:typical_cpu_gpu_tops}.

\setcounter{figure}{0}
\setcounter{table}{0}

\begin{table}[h]
\centering
\begin{tabular}{l|l|l|l|l|l|l|l|l}
CPU/GPU                & GHz & Watts  & fp32   & fp16      & int8   \\
\hline
ARM Cortex A-7\phantom{5}(2-core)& 1.5   & 1.5  &  3      & $\dagger$ & 12     \\
ARM Cortex A-75(4-core)& 3.0   & 4    &  48     & 96        & 96    \\
NVIDIA Turing Tesla T4 & 1.35  & 70   &  8100   & 16200     & 130000\\ 
\end{tabular}
\caption{Peak GFlops/Gops according to specification}
\label{tab:typical_cpu_gpu_tops}
\end{table}

In order to decrease the storage and compute requirements of the model during inference its parameters are often stored as integers with a low number of bits. It is common to use 8-bit integers (1 Byte), rather than 16- (2 Bytes) or 32-bit (4 Bytes) floating point numbers. The process of converting model parameters from ``continuous" floating point to discrete integer numbers is called quantization.

Let the original optimization problem be
\begin{equation}
\min_w L(w,x)
\label{eq:orig}
\end{equation}
where $L$ is the loss measured during training. There are many different quantization schemes based on symmetric vs. asymetric intervals, uniform vs. non-uniform discrete partitioning, different rounding modes and choices for handling the outliers, e.g. few elements that lie outside of the range of most of the other elements. 

Let us consider a uniform quantization of the parameter weights $W = [w_{ij}]$, for example from the either convolution \eqref{eq:convolution} or fully connected layers \eqref{eq:fully_connected}, and activations. Then, the quantized problem is commonly written as
\begin{equation}
\min_w L(Q(w),Q(x))
\label{eq:quant}
\end{equation}
where $Q$ is the quantization function.

For instance, if we use only two intervals then the process is referred to as binarization and resulting element can be stored in a single bit \cite{Courbariaux2015,Courbariaux2016,Hubara2016,Rastegari2016}. It can be performed using a single threshold point $\epsilon$ as shown below
\begin{equation}
Q(w) = \left\{ \begin{array}{l} -1 \texttt{ if } w < \epsilon \\ +1 \texttt{ otherwise }  \end{array} \right.
\label{eq:binary}
\end{equation}

The ternary networks use three intervals with resulting elements stored in 2 bits \cite{Li2016,Mellempudi2017,Choi2018}. Then, quantization can be performed using two threshold points $\epsilon_1$ and $\epsilon_2$ resulting in
\begin{equation}
Q(w) = \left\{ \begin{array}{l} -1 \texttt{ if } w\phantom{|} < \epsilon_1 \\ \phantom{-}0 \texttt{ if } \epsilon_1 \ge w \ge \epsilon_2 \\ +1 \texttt{ if } w\phantom{|} > \epsilon_2 \end{array} \right.
\label{eq:ternary}
\end{equation}

Finally, let arbitrary \# of bits $t$ correspond to $2^t$ points and $2^t+1$ intervals. Let us assume that we would like to quantize floating point number $w \in [a,b]$, with $c = \max(|a|,|b|)$ and length $d = (b - a)/2$. A uniform quantization can be performed symetrically in the interval $[-c,c]$ using multiplier $\gamma  = c/(2^{t-1}-1)$, so that
\begin{equation}
Q(w) = \gamma * \texttt{round}( w / \gamma ) 
\label{eq:arbitrary}
\end{equation}
with $2^t-1$ effective points because $0.0$ is double counted.

On the other hand, notice that we can shift the interval $[a,b]$ to the interval $[-d/2,d/2]$ located around $0.0$ by adding a scalar bias term $s = (a+b)/2$. Therefore, uniform asymetric quantization can be performed with bias $s$ using multiplier $\delta  = d/(2^{t}-2)$, so that
\begin{equation}
Q(w) = \delta * \texttt{round}( (w-s)/ \delta ) + s
\label{eq:arbitrary_shift}
\end{equation}
where \texttt{round} operation rounds a floating point to an integer value \cite{Wen2016,Jacob2017,Krishnamoorthi2018}. 

The advantage of symmetric quantization is that for sparse parameters, with a lot of $0.0$ elements, the sparsity is preserved. Note that computation with zeroes can be skipped in hardware \cite{Albericio2016, Venkatesh2016, Reagan2016, Chen2017, Kim2017, Parashar2017}. The disadvatage is that for highly asymetric intervals many discrete representations may be wasted.

The non-uniform quantization assigns discrete points to the interval based on the distribution of floating point values in it \cite{Bagherinezhad2017,Wang2018}. Therefore, it does not have a fixed stride from one point to the next. Its advantage is that the encoded values are more representative of the original ones, but at the same time it can be hard to map back and perform operations with them.     

The techniques for handling outliers and determining maximum thresholds, e.g. using adaptive schemes, Kullback–Leibler (KL) divergence measured loss of information, or L2 error minimization in Caffe2, have been investigated in \cite{jia2014caffe,Migasz2017,Zhou2017,Park2018}.  

However, independent of all of these choices, notice that a common trend among \eqref{eq:binary} - \eqref{eq:arbitrary_shift} is that fixed thresholds are used in quantization function $Q$ to clamp floating point values to discrete points. We point out that the matrix- and neural network-based compression techniques are outside the scope of this paper \cite{Gong2014,Denton2014,Jaderberg2014,Mishra2018}.

In this paper we will focus on a very different approach for uniform quantization using periodic functions, such as trigonometric sine (or cosine) as well as hat functions. We discuss uniform quantization, but our ideas can be generalized to non-uniform case using variations of these periodic functions with decaying amplitude and increasing base lengths away from the origin \cite{Stenger1993,Strang2008}. 

\section{Periodic Functions as Regularizers} 

We propose an unorthodox approach for quantizing the weights of a neural network. Instead of using quantization function $Q$, we propose proposed adding a regularization term $R$ to the loss, so that the resulting optimization problem is written as
\begin{equation}
\min_w L(w,x) + \lambda R(w,x) 
\label{eq:orig_reg}
\end{equation}
where $\lambda$ is a scalar scaling parameter.

The regularization term $R$ is a sum of periodic functions that push the values of the weights (and potentially activations) to a set of discrete points during training. Next we will discuss different choices for these functions. 

\subsection{Trigonometric (Continuous) Functions}
Let us focus only on the weights and use trigonometric sine, so that 
\begin{equation}
R(w) = \sum_{\forall w} \text{amplitude} * \sin^2 (\pi * \text{frequency} * (w/c)) 
\label{eq:sin_con}
\end{equation}
where $c$ is the maximum weight in absolute value as defined in \eqref{eq:arbitrary}.

\begin{figure}[h]
 \begin{center}
  \begin{subfigure}[b]{0.49\textwidth} 
   \includegraphics[width=\textwidth]{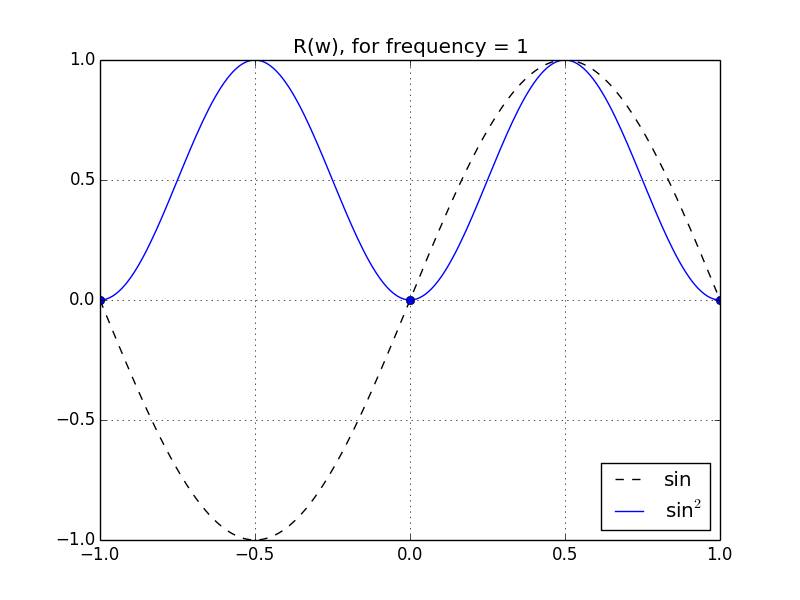}
   \caption{frequency=1}
   \label{fig:sin3}
  \end{subfigure} 
  \begin{subfigure}[b]{0.49\textwidth}
   \includegraphics[width=\textwidth]{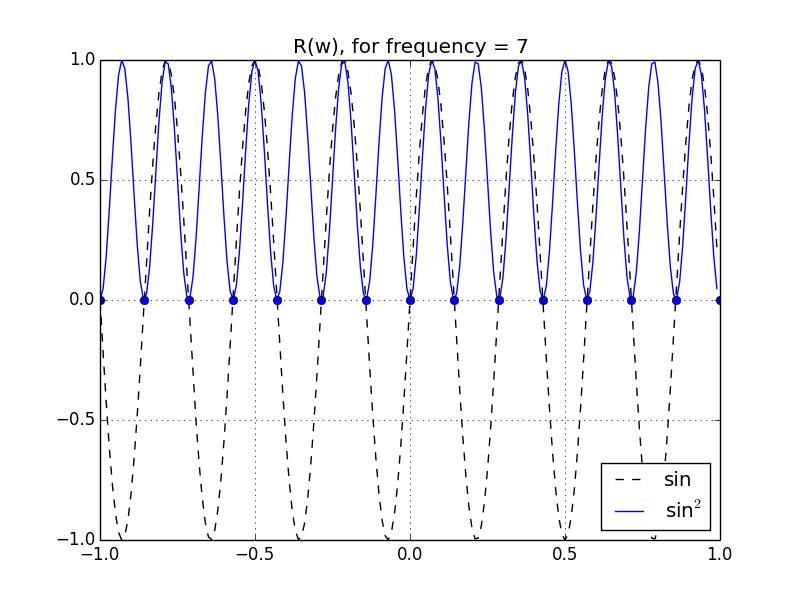}
   \caption{frequency=7}
   \label{fig:sin15}
  \end{subfigure}
  \caption{Plot of sine and its square for amplitude=1}
  \squeeze \squeeze
 \end{center}
\end{figure}

Notice that for frequency=1 the function R(w) attains its minimum 0.0, when the weight values are distributed at 3 discrete locations, while for frequency=7 it attains its minimum 0.0, when the weight values are distributed at 15 discrete locations, as shown on Fig. \ref{fig:sin3} and \ref{fig:sin15}, respectively.

\begin{figure}[h]
 \begin{center}
  \begin{subfigure}[b]{0.49\textwidth}
   \includegraphics[width=\textwidth]{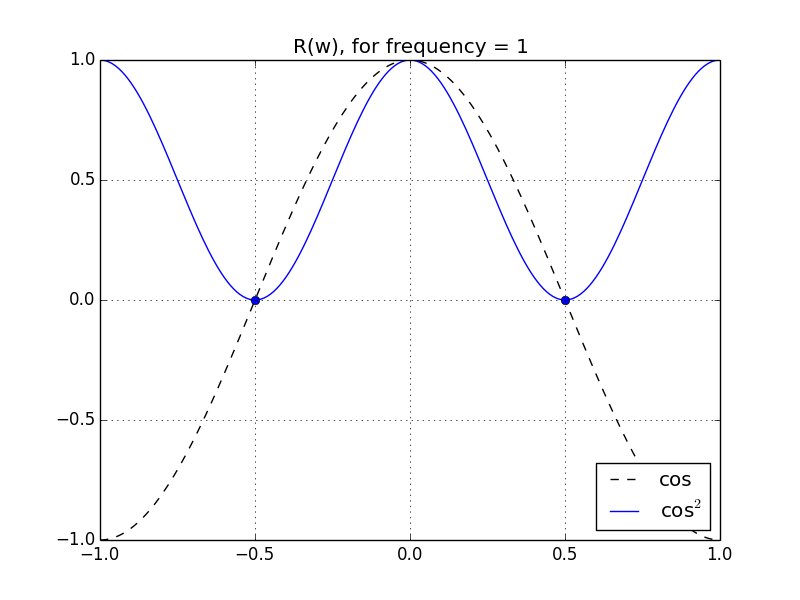}
   \caption{frequency=1}
   \label{fig:cos2}
  \end{subfigure}
  \begin{subfigure}[b]{0.49\textwidth}
   \includegraphics[width=\textwidth]{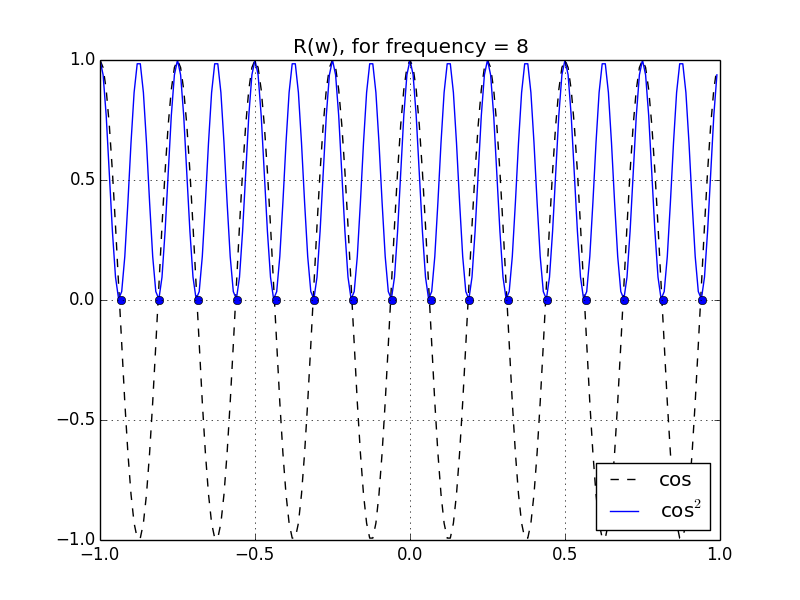}
   \caption{frequency=8}
   \label{fig:cos16}
  \end{subfigure}
  \caption{Plot of cosine and its square for amplitude=1}
  \squeeze \squeeze
 \end{center}
\end{figure}

\clearpage
\newpage

An analogous plot can also be created using a trigonometric cosine function and its square, with a difference being a shift in the discrete points so that there are 16 rather than 15 of the points where minimum is attained, but perhaps most importabtly $0.0$ is not preserved, see Fig, \ref{fig:cos2} and \ref{fig:cos16}.

\subsection{Hat (Non-Continuous) Functions}

Let us focus only on the weights and use a hat function, so that 
\begin{equation}
R(w) = \sum_{\forall w} \text{amplitude} * |((\text{frequency} * ((w/c)-0.5)) \% 1)*2 - 1| 
\label{eq:hat_non}
\end{equation}
where $c$ is the maximum weight in absolute value as defined in \eqref{eq:arbitrary}.

\begin{figure}[h]
 \begin{center}
  \begin{subfigure}[b]{0.49\textwidth}
    \includegraphics[width=\textwidth]{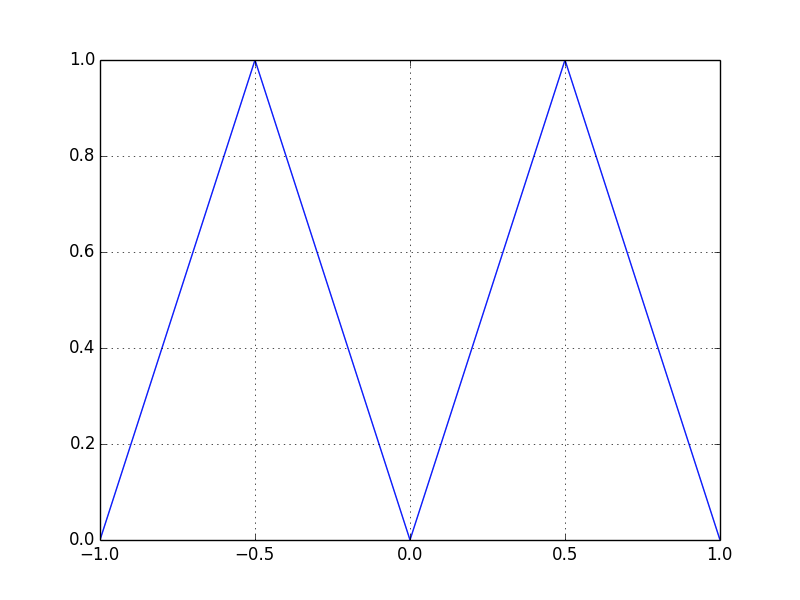}
    \caption{frequency=1}
    \label{fig:hat3}
  \end{subfigure}
  \begin{subfigure}[b]{0.49\textwidth}
    \includegraphics[width=\textwidth]{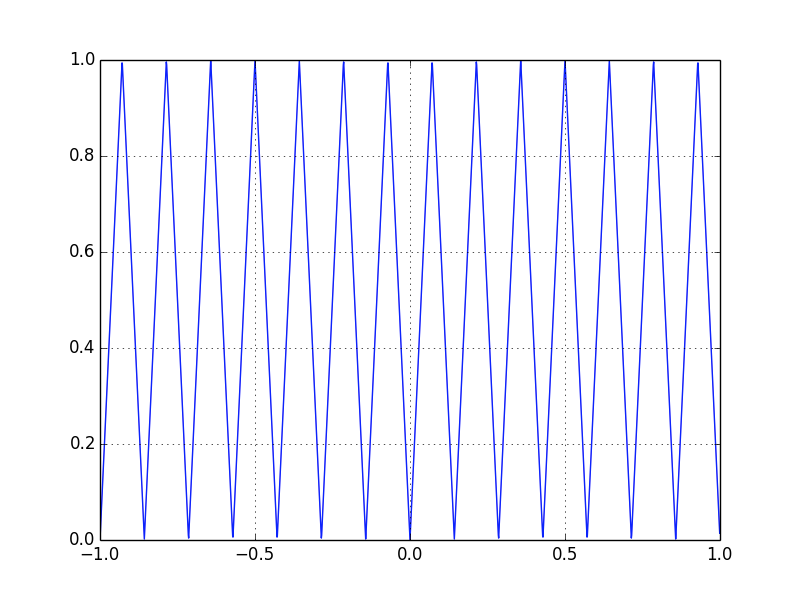}
    \caption{frequency=7}
    \label{fig:hat15}
  \end{subfigure}  
  \caption{Plot of hat for amplitude=1}
  \squeeze \squeeze
 \end{center}
\end{figure}

Once again, notice that for frequency=1 the function R(w) attains its minimum 0.0, when the weight values are distributed at 3 discrete locations, while for frequency=7 it attains its minimum 0.0, when the weight values are distributed at 15 discrete locations, as shown on Fig. \ref{fig:hat3} and \ref{fig:hat15}, respectively. Here we show variant of the hat function corresponding to sine, while a shifted variant corresponding to cosine is also possible. 

Notice that the use of regularization for the purpose of quantization has been suggested in \cite{Hung2015}. However, in this earlier work the authors use distance from fixed points (centroids) $q$ as a penalty measure to ensure quantization. This contrasts with our periodic trigonometric sine and hat functions, with amplitude and frequency hyper-parameters defined in \eqref{eq:sin_con} and \eqref{eq:hat_non}, respectively. 

It is important to highlight a few differences between sine (or cosine) and hat functions. Notice that sine function has very nice properties. It is periodic, continuous and differentiable. However, it is not convex, unlike many of the existing regularizers. Also, notice that the maximum value of the regularizer R(w) is know ahead of time. It can be computed by assuming that all weights translate into value 1.0 after application of the $\sin^2$ function. Then, the regularizer can be scaled by a constant $\gamma$, such that $\gamma$R(w) $\in [0,1]$. This can be used to facilitate and in fact define the regularizer scaling $\lambda$ in \eqref{eq:orig_reg}, therefore reducing the number of hyper-parameters.

Also, sine function has a gradient that is zero (or close to zero) in the neighborhood of points where it attains it's minimum and maximum values. This property might make escaping the maximum or approaching the minimum slow in their respective neighborhoods. On the other hand, hat function is non-convex and non-continuous, with constant gradient towards the minimum except for the points where it attains its minimum and maximum values, where the gradient does not exist. These tradeoffs might guide the choice between these functions, in a way similar to that of a choice between Sigmoid and ReLU activation functions.
 
Finally, notice that amplitude can be changed adaptively during the training procedure, which allows us to obtain higher test accuracy, as will be shown in the experiments section. The frequency can also be varied during training, but these experiments are outside of the scope of this paper.  

\subsection{From Bits to Frequency and Vice-Versa} 
In practice we are interested in selecting the number of bits $t$ to be used for quantization. For the sine and associated hat function the frequency corresponding to $t>1$ number of bits can be found by using
\begin{equation}
\text{frequency} =  2^{t - 1} - 1
\end{equation}
and vice-versa
\begin{equation}
t = \lceil \log_2 (\text{frequency} + 1) + 1 \rceil
\end{equation}
so that frequency 1 implies 2 bits, while frequency 7 implies 4 bits.

On the other hand, for cosine the frequency corresponding to $t>0$ number of bits can be found by using 
\begin{equation}
\text{frequency} =  2^{t-1}
\end{equation}
and vice-versa
\begin{equation}
t = \lceil \log_2 (\text{frequency}) + 1 \rceil
\end{equation}

For instance, frequency 1 implies 1 bit, while frequency 8 implies 4 bits, and so on and so forth.

\section{Experiments}

In this section we will investigate the accuracy of ResNet-20 on CIFAR-10 and ResNet-50 on ImageNet datasets \cite{ResNet,CIFAR,ImageNet}. We will compare the test error achieved by the original and quantized models with loss function defined in \eqref{eq:orig} and \eqref{eq:orig_reg}, respectively. The regularization term we add to the loss in \eqref{eq:orig_reg} relies on periodic functions: trigonometric sine in \eqref{eq:sin_con} and hat in \eqref{eq:hat_non}. It can be computed using the following PyTorch \cite{PyTorch} code snippet

\begin{verbatim}
def periodic_regularization(model, amplitude, frequency):
    pi = 3.141592
    total = 0
    for m in model.modules():
        if isinstance(m, nn.Conv2d) or isinstance(m, nn.Linear):
           ic = 1/w.abs().max()
           rw = torch.sum(amplitude * 
             #either sin
             torch.pow(torch.sin(pi * frequency * (w * ic)), 2))
             #or hat function
             torch.abs(((((w * ic) - 0.5) * frequency) % 1) * 2 -1))
\end{verbatim}

The training is performed using batch size 256 with default 100 epochs for CIFAR-10 and 90 epochs for ImageNet dataset. We use a fixed schedule that adjusts the amplitude hyper-parameter every $30$ epochs. We start with a small amplitude, such as $10^{-4}$, and progressively adjust it until it reaches, say $10^{-1}$ after typical $100$ epochs of training. Notice that amplitude subsumes the scaling hyper-parameter $\lambda$, which is always set to 1.0. Note that other than using a fixed schedule we do not require any special treatment for the first or last model layers or training epochs, which is otherwise often required to produce good approximations. We show the results of representative  runs.

After training the model is quantized using symmetric uniform quantization in \eqref{eq:arbitrary}, which can be performed using the following PyTorch \cite{PyTorch} code snippet

\begin{verbatim}
def quantize_model(model, frequency):
    def quantize_weights(m):
        if isinstance(m, nn.Conv2d) or isinstance(m, nn.Linear):
            c = m.weight.abs().max().data
            m.weight.data.mul_(frequency/c)
            m.weight.data.round_()
            m.weight.data.mul_(c/frequency)

    model.apply(quantize_weights)
\end{verbatim}

\begin{figure}[h]
 \begin{center}
  \includegraphics[width=8cm]{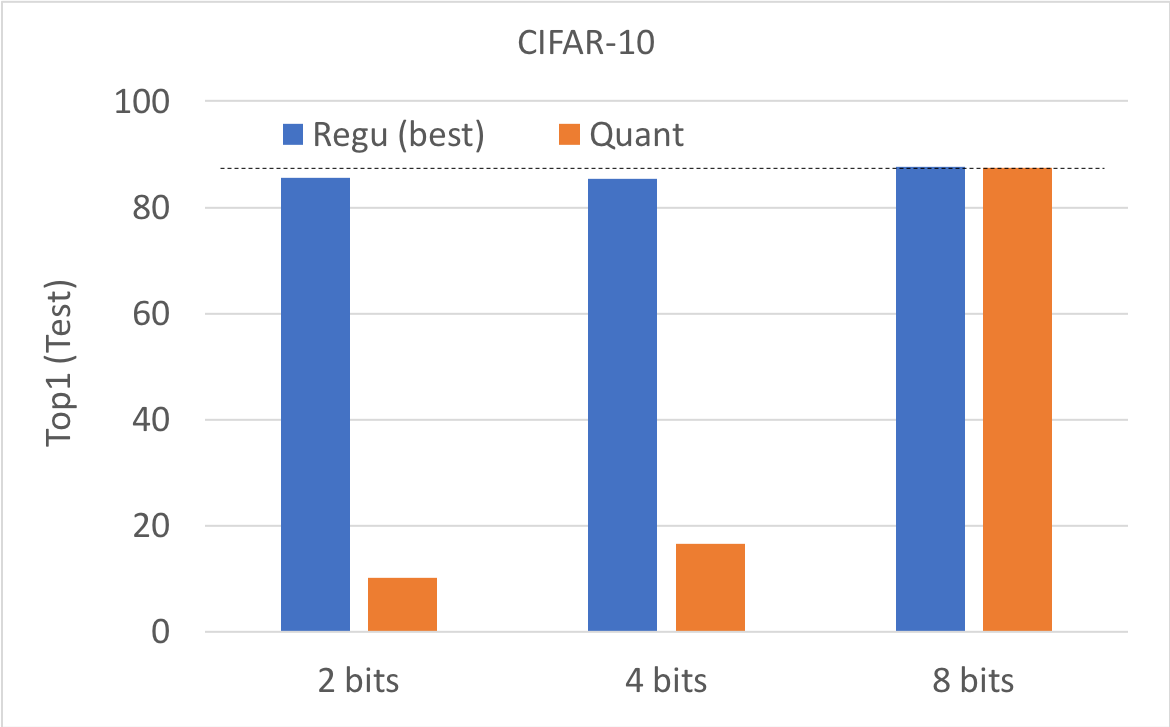}
  \caption{Plot of Top1 test error for CIFAR-10 dataset with Resnet20}
  \label{fig:cifar10_resnet20_variable_amp}
  \squeeze 
 \end{center}
\end{figure}

We illustrate the difference between original, original with regularization (Regu), and quantized model (Quant) for CIFAR-10 dataset on Fig. \ref{fig:cifar10_resnet20_variable_amp}. The accuracy of the original model is plotted with a black dotted line, while the accuracy of other models is plotted with color bars. Notice that the model accuracy changes significantly depending on the number of bits used for quantization. For instance, there seems to be a clear boundary between 4 and 8 bits, where there seems to be (not or) enough bits to represent the information. Notice that while the training succeeds in all cases, the quantization fails to produce accurate results with less than 8 bits.

\begin{figure}[h]
 \begin{center}
  \includegraphics[width=8cm]{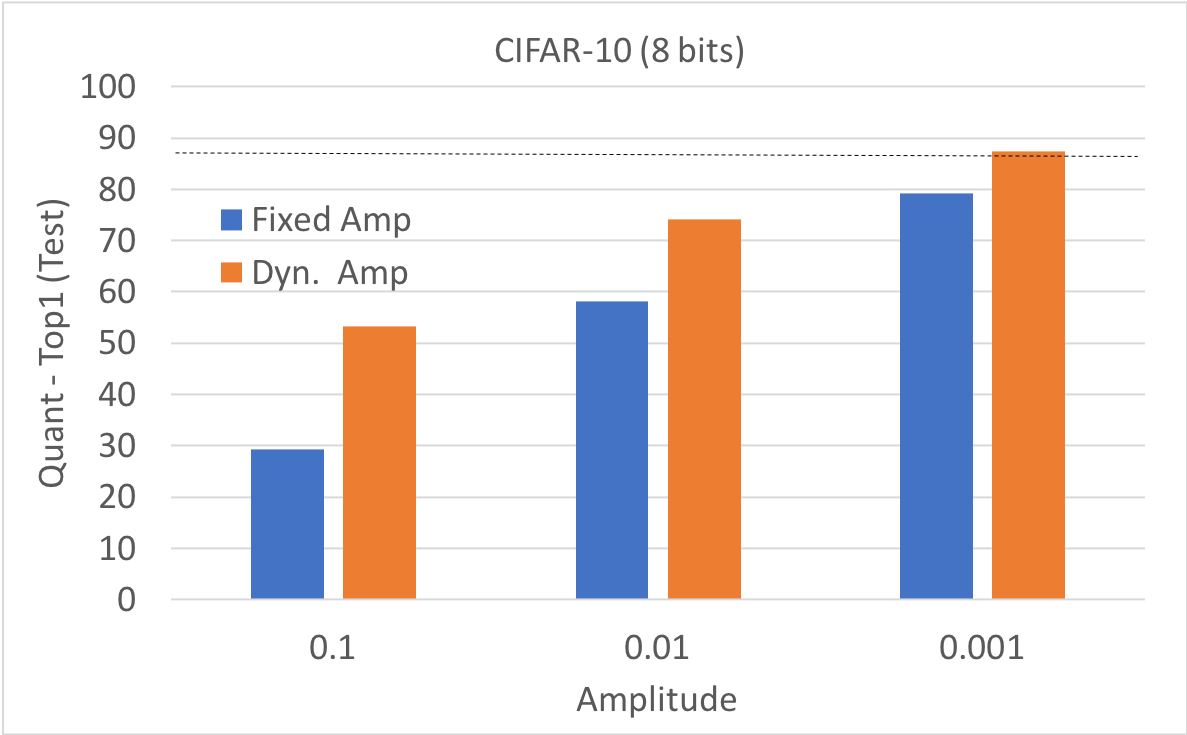}
  \caption{Plot of Top1 test error for CIFAR-10 dataset with Resnet-20}
  \label{fig:cifar10_resnet20_fixed_amp}
  \squeeze 
 \end{center}
\end{figure}

\begin{figure}[h]
 \begin{center}
  \includegraphics[width=8cm]{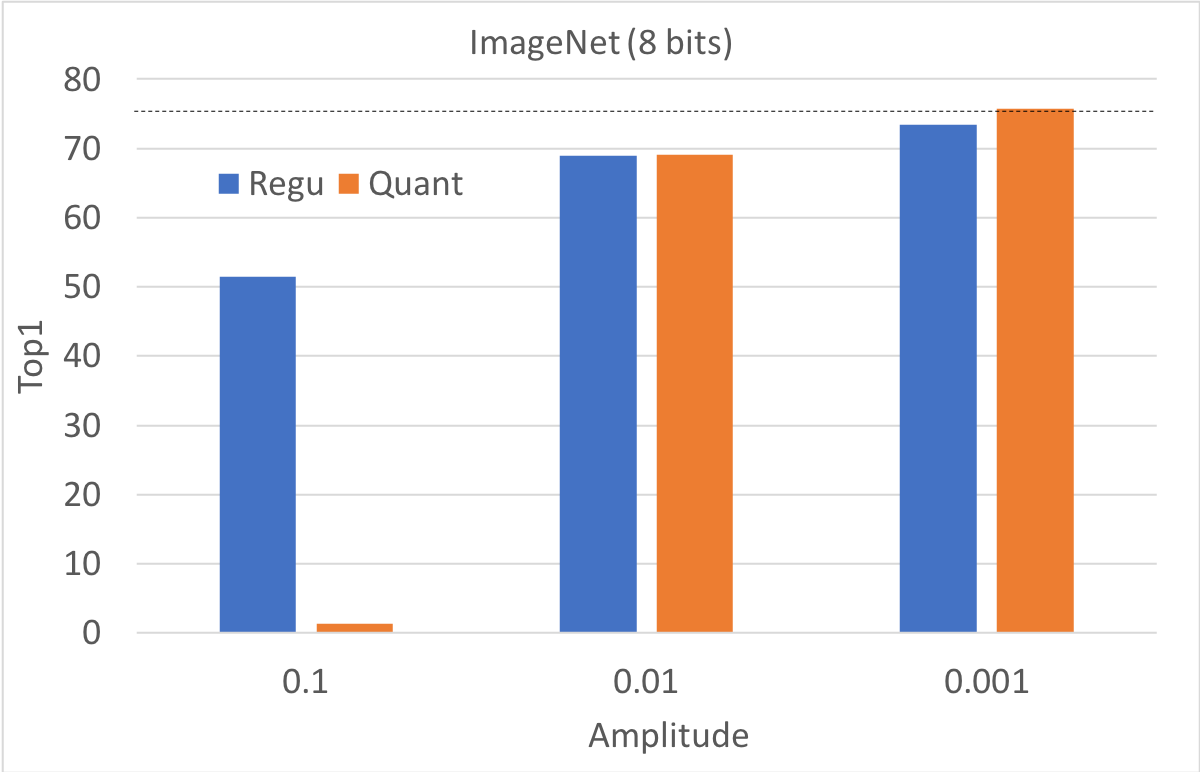}
  \caption{Plot of Top1 test error for ImageNet dataset with Resnet-50}
  \label{fig:imagenet_resnet50}
  \squeeze 
 \end{center}
\end{figure}

Also, we illustrate the attained model accuracy with different starting amplitudes for CIFAR-10 dataset on Fig. \ref{fig:cifar10_resnet20_fixed_amp}. The accuracy of the original model is plotted with a black dotted line, while the accuracy of the 8-bit quantized model is plotted with color bars. Notice that using adaptive rather than static amplitude allows us to reach higher test accuracy. Also, in our experiments we have found that it is a good practice to target the initial amplitude and choice of fixed schedule such that the final amplitude is in the range of 0.01 - 0.001, which would correspond to a reasonable value of the regularization scaling $\lambda$. We observe similar results on the ImageNet dataset, as seen on Fig. \ref{fig:imagenet_resnet50}. 

\begin{figure}[h]
 \begin{center}
  \includegraphics[width=8cm]{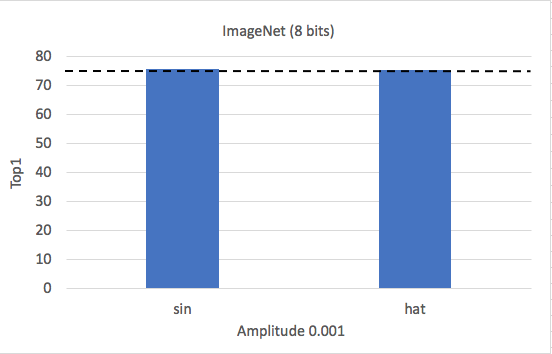}
  \caption{Plot of Top1 test error for ImageNet dataset with Resnet-50}
  \label{fig:imagenet_hat_resnet50}
  \squeeze 
 \end{center}
\end{figure}

Finally, notice that both sine and hat functions perform as well on the ImageNet dataset, as shown in Fig. \ref{fig:imagenet_hat_resnet50}. Once again, the accuracy of the original model is plotted with a black dotted line, while the accuracy of the 8-bit quantized model is plotted with color bars. In all plots, amplitude denotes the final amplitude. The detailed results are also summarized in tables Tab. \ref{tab:cifar10_summary} and \ref{tab:imagenet_summary}.

\begin{table}[h]
\centering
\begin{tabular}{l|c|c|c|c|c|c|c|c}
   & Default Model & \multicolumn{5}{c|}{Quantized model (with sine)} \\
\hline             
           & test (best)   & \multicolumn{3}{c|}{8 bits}   & 4 bits & 2 bits \\
\hline
Amplitude  & n/a           & 0.1   & 0.01  & 0.001         & 0.001 & 0.001 \\
\hline
Test error (fixed) & 84.72 (87.70) & 29.26 & 58.18 & 79.18 & n/a   & n/a   \\
Test error (dyn)   & 84.72 (87.70) & 53.28 & 74.14 & 87.46 & 16.66 & 10.20 \\
\end{tabular}
\caption{CIFAR-10: summary of representative experiments}
\label{tab:cifar10_summary}
\squeeze 
\end{table}

\begin{table}[h]
\centering
\begin{tabular}{l|c|c|c|c|c|}
          & Default Model & \multicolumn{3}{c|}{Quantized model (with sine)} & (with hat) \\
\hline             
                &         & \multicolumn{3}{c|}{8 bits} & 8 bits \\
\hline
Amplitude       &  n/a    & 0.1   & 0.01  & 0.001   & 0.001 \\
\hline
Top1 error      & 75.84   & 1.29 & 69.02 & 75.77   & 75.57  \\
Top5 error      & 92.90   & 4.62 & 89.27 & 92.54   & 92.58  \\
\end{tabular}
\caption{ImageNet: summary of representative experiments}
\label{tab:imagenet_summary}
\squeeze
\end{table}

\section{Conclusion and Future Work}

We have proposed a novel technique for quantizing neural networks, based on regularization with periodic functions. We have shown that it can be effectively used to quantize ResNets on CIFAR-10 and ImageNet datasets. In our experiments we have achieved virtually no losses vis-à-vis standard model by using amplitude scaling on a fixed schedule through training followed by 8-bit integer quantization. While similar quality results exist for quantization of CNNs, in this note we have achieved them through a completely novel method. In the future, we would like to incorporate the quantization of activations into this approach and experiment with more classes of neural networks.

\section*{Acknowledgements}

The authors would like to thank Marat Dukhan, Bram Wasti and Satish Nadathur for collecting ARM Cortex A-7 and A-75 processor intruction information as well as Misha Smelyanskiy for his helpful comments and suggestions.

\bibliography{quantization_using_periodic_functions}
\bibliographystyle{iclr2018_conference}

\clearpage
\newpage

\section{Appendix: Brief Background}

The machine learning models are used in the fields of computer vision (CV) and natural language processing (NLP) among many others. In particular, the deep learning models based on neural networks composed of multiple layers have achieved unprecedented gains in accuracy of image classification and object detection tasks \cite{Krizhevsky2012,Szegedy2014,MNIST,CIFAR,ImageNet}. 

In this paper we focus on the CV deep learning models that often rely on convolutional neural networks (CNNs), that are mainly composed of multiple convolution, fully connected and batch normalization layers \cite{LeNet1989a,LeNet1989d,Goodfellow2016,Ioffe2015}. For example, we will investigate ResNet-20 on CIFAR-10 and ResNet-50 on ImageNet datasets \cite{ResNet,CIFAR,ImageNet}. For completeness we review the most common layers next.

The convolution layer can be defined as 
\begin{equation}
\left\{
\begin{array}{l}
Z = f(Y) \\
Y = X \odot W
\end{array}
\right.
\label{eq:convolution}
\end{equation}
where input image $X \in \mathbb{R}^{h \times w \times c}$, the filter $W \in \mathbb{R}^{m \times n \times c}$, while $\odot$ denotes a convolution\footnote{In this context it is also common to use a cross-correlation rather than a convolution.} and $f$ denotes a pooling operation, resulting in output $Z \in \mathbb{R}^{h' \times w'}$ with $h'=(h-m)/s_1+1$ and $w'=(w-n)/s_2+1$ for strides $s_1,s_2 \ge 1$. The operation is usually repeated for $c'$ output channels, resulting in $Z \in \mathbb{R}^{h' \times w' \times c'}$.  

The fully connected layer is defined as
\begin{equation}
\left\{
\begin{array}{l}
Z = f(Y) \\
Y = WX + \textbf{b}\textbf{e}^{T}
\end{array}
\right.
\label{eq:fully_connected}
\end{equation}
where input $X = [\textbf{x}_1,...,\textbf{x}_r] \in \mathbb{R}^{n \times r}$, weights $W \in \mathbb{R}^{m \times n}$, bias $\textbf{b} \in \mathbb{R}^n$, unit vector $\textbf{e}^{T}=[1,...,1]$, the non-linear activation function $f$ is applied component-wise on intermediate $Y \in \mathbb{R}^{m \times r}$ and output $Z\in \mathbb{R}^{m \times r}$ for $r \ge 1$ batch size. 

The typical batch normalization layer can be written as
\begin{equation}
\left\{
\begin{array}{l}
Z = WY +\textbf{b}\textbf{e}^{T} \\
Y = \frac{1}{\sqrt{r}}X(I - \frac{1}{r}\textbf{e}\textbf{e}^T)
\end{array}
\right.
\label{eq:normalization}
\end{equation}
where input $X = [\textbf{x}_1,...,\textbf{x}_r] \in \mathbb{R}^{n \times r}$, scaled diagonal matrix of weights $W \in \mathbb{R}^{m \times m}$, bias $\textbf{b} \in \mathbb{R}^n$, unit vector $\textbf{e}^{T}=[1,...,1]$ for $r \ge 1$ batch size \cite{Devarakonda2017}.

\end{document}